\documentclass{article}

\usepackage{PRIMEarxiv}

\usepackage[utf8]{inputenc} 
\usepackage[T1]{fontenc}    
\usepackage{hyperref}       
\usepackage{url}            
\usepackage{booktabs}       
\usepackage{amsfonts}       
\usepackage{nicefrac}       
\usepackage{microtype}      
\usepackage{lipsum}
\usepackage{fancyhdr}       
\usepackage{graphicx}       
\graphicspath{{media/}}     
\usepackage[section]{placeins}
\begin{document}
  
\title{Slice-Wise Quality Assessment of High b-Value Breast DWI via Deep Learning–Based Artifact Detection}
\maketitle

{\large
\textbf{Ameya Markale\textsuperscript{*,1,2},
Luise Brock\textsuperscript{*,1,3},
Ihor Horishnyi\textsuperscript{1},
Dominika Skwierawska\textsuperscript{1},
Tri-Thien Nguyen\textsuperscript{1,4},
Hannes Schreiter\textsuperscript{1},
Shirin Heidarikahkesh\textsuperscript{1},
Lorenz A. Kapsner\textsuperscript{1,5},
Michael Uder\textsuperscript{1},
Sabine Ohlmeyer\textsuperscript{1},
Frederik B Laun\textsuperscript{1},
Andrzej Liebert\textsuperscript{1,2}, and
Sebastian Bickelhaupt\textsuperscript{1}
}}

\vspace{2mm}

1.\ Institute of Radiology, Universitätsklinikum Erlangen, Friedrich-Alexander-Universität Erlangen-Nürnberg (FAU), Maximiliansplatz 3, 91054 Erlangen, Germany\\
2.\ Institute of Computer Science, Polish Academy of Science, Warsaw, Poland\\
3.\ Department of Artificial Intelligence in Biomedical Engineering (AIBE), Friedrich-Alexander-Universität Erlangen- \ Nürnberg (FAU), Erlangen, Germany\\
4. Pattern Recognition Lab, Friedrich-Alexander-Universität Erlangen-Nürnberg (FAU), Erlangen, Germany\\
5. Medical Informatics, Friedrich-Alexander-Universität Erlangen-Nürnberg (FAU), Erlangen, Germany

Corresponding Author: Ameya Markale\\
E-Mail: ameya.markale@ipipan.waw.pl\\
\textsuperscript{*}A.M. and L.B. contributed equally to the work.\\

\begin{abstract}
Diffusion-weighted imaging (DWI) can support lesion detection and characterization in breast magnetic resonance imaging (MRI), however especially high b-value diffusion-weighted acquisitions can be prone to intensity artifacts that can affect diagnostic image assessment. This study aims to detect both hyper- and hypointense artifacts on high b-value diffusion-weighted images (b = 1500~s/mm$^2$) using deep learning, employing either a binary classification (artifact presence) or a multiclass classification (artifact intensity) approach on a slice-wise dataset. 
This IRB-approved retrospective study used the single-center dataset comprising n = 11806 slices from routine 3T breast MRI examinations performed between 2022 and mid-2023. Three convolutional neural network (CNN) architectures (DenseNet121, ResNet18, and SEResNet50) were trained for binary classification of hyper- and hypointense artifacts. The best performing model (DenseNet121) was applied to an independent holdout test set and was further trained separately for multiclass classification. Evaluation included area under receiver operating characteristic curve (AUROC), area under precision recall curve (AUPRC), precision, and recall, as well as analysis of predicted bounding box positions, derived from the network’s Grad-CAM heatmaps. 
DenseNet121 achieved AUROCs of 0.92 and 0.94 for hyper- and hypointense artifact detection, respectively, and weighted AUROCs of 0.85 and 0.88 for multiclass classification on single-slice high b-value diffusion-weighted images. A radiologist evaluated bounding box precision on a 1--5 Likert-like scale across 200 slices, achieving mean scores of 3.33 ± 1.04 for hyperintense artifacts and 2.62 ± 0.81 for hypointense artifacts.
Hyper- and hypointense artifact detection in slice-wise breast DWI MRI dataset (b = 1500~s/mm$^2$) using CNNs particularly DenseNet121, seems promising and requires further validation.
\end{abstract}

\bigskip

\newpage

\section{Introduction}
Multiparametric imaging protocols in breast magnetic resonance imaging (MRI) increasingly incorporate diffusion weighted imaging (DWI) sequences \cite{le2025fat}. DWI is of interest as it enables both the visual depiction and the (semi-) quantitative assessment of lesions, supporting detection and characterization of lesions \cite{le2025fat, partridge2017diffusion, christner2024breast}. It provides microstructural information by measuring the Brownian motion of water molecules and thereby characterizing water diffusivity \cite{le2025fat, kapsner2023image}. 

Additionally, due to the capabilities of DWI to map distinct and possibly complementary microstructural properties, DWI is increasingly used and investigated in studies including breast imaging applications \cite{le2025fat}.
Whilst DWI has been suggested to potentially complement diagnostic assessment or to even serve as a non-contrast enhanced stand-alone tool, it remains technically challenging and prone to various image artifacts. Image artifacts in DWI can originate from various sources involving the scanner, acquisition techniques and the patient, which commonly result in two major categories of visual artifact impression: either hyperintense or hypointense artifacts. Sources for hyperintense artifacts on diffusion-weighted images can, amongst others, be related to skin folds and failed fat suppression. Hypointense artifacts can, amongst others, be related to pulsation artifacts or susceptibility related signal loss \cite{le2025fat, peters2008meta, partridge2017dwi}.

Detecting and classifying the artifacts in DWI is of diagnostic and technical relevance, as artifacts can obscure or mimic pathologies and impede the diagnostic assessment or lead to miscalculations in subsequent quantification maps using the apparent diffusion coefficient. Artifact detection is also important for Artificial Intelligence applications, as artifacts can bias model training, degrade inference performance, cause incorrect segmentation and reduce the reliability of lesion detection or virtual contrast enhancement generation especially when diffusion-weighted images are used as inputs \cite{schreiter2024virtual, westfechtel2025ai, liebert2025impact, herent2019detection, maicas2017deep}.

Previously, Kapsner et al. used convolutional neural networks (CNNs) to detect artifacts in high b-value DWI maximum intensity projections (MIPs) of the breast. This study was conducted using MIPs only, where artifacts originating from a limited portion of the volume can extend across the entire projection, with potential implications for image interpretation.

We now propose a deep learning-based approach using CNNs on a slice-based level to detect hyper- and hypointense artifacts using either binary classification or multiclass classification network to grade the artifact severity in high b-value DWI \cite{o2015introduction}. Grading artifact severity is important because different artifact strengths have different clinical or technical implications. Minor artifacts may be acceptable for diagnostic evaluation, but moderate and severe artifacts can hamper diagnostic evaluation.

\section{Materials and Methods}
\label{sec:headings}
\subsection{Datasets}
This retrospective IRB-approved study included a dataset of n = 1383 routine breast MRI acquisitions acquired between 2022 and mid-2023. Imaging was performed on 3T MRI systems (Magnetom Skyra Fit and Magnetom Vida, Siemens Healthineers, Erlangen, Germany) using an 18-channel breast coil (Siemens Healthineers, Erlangen, Germany) and a multiparametric protocol, which includes T1-weighted (before and dynamically after contrast agent application), T2-weighted fat saturated and multi b-value DWI acquisitions. Patients (mean age: 48 ± 12 years) were scanned in prone position with both arms placed alongside the body. Examinations of patients with breast implants were excluded, as DWI acquisition can be limited due to echo planar imaging (EPI) artifacts in silicone \cite{rotili2023unenhanced}.

DWI was performed using a single shot spin echo EPI sequence with multiple b-values (b = 50, 750, 1500~s/mm$^2$). For this study, we focused specifically on the acquisitions using high b-value (b = 1500~s/mm$^2$) as high b-values are reported to be especially relevant for lesion detection due to the signal decay in healthy fibroglandular tissue and simultaneously can be more prone to developing artifacts \cite{ohlmeyer2021ultra, bickel2019diffusion}. Detailed acquisition parameters for the DWI sequences are summarized in Supplemental Table \ref{tab:ST1}. The data has previously been investigated and published, especially with regards to artifacts in DWI, however only investigating hyperintense artifacts on MIPs. This work now investigates the feasibility and value of dual detection of hyper- and hypointense artifacts on individual single slices and for different artifact-classes.

\subsection{Experimental Setup Overview}
\begin{figure}[h!]
\centering
\includegraphics[scale=1]{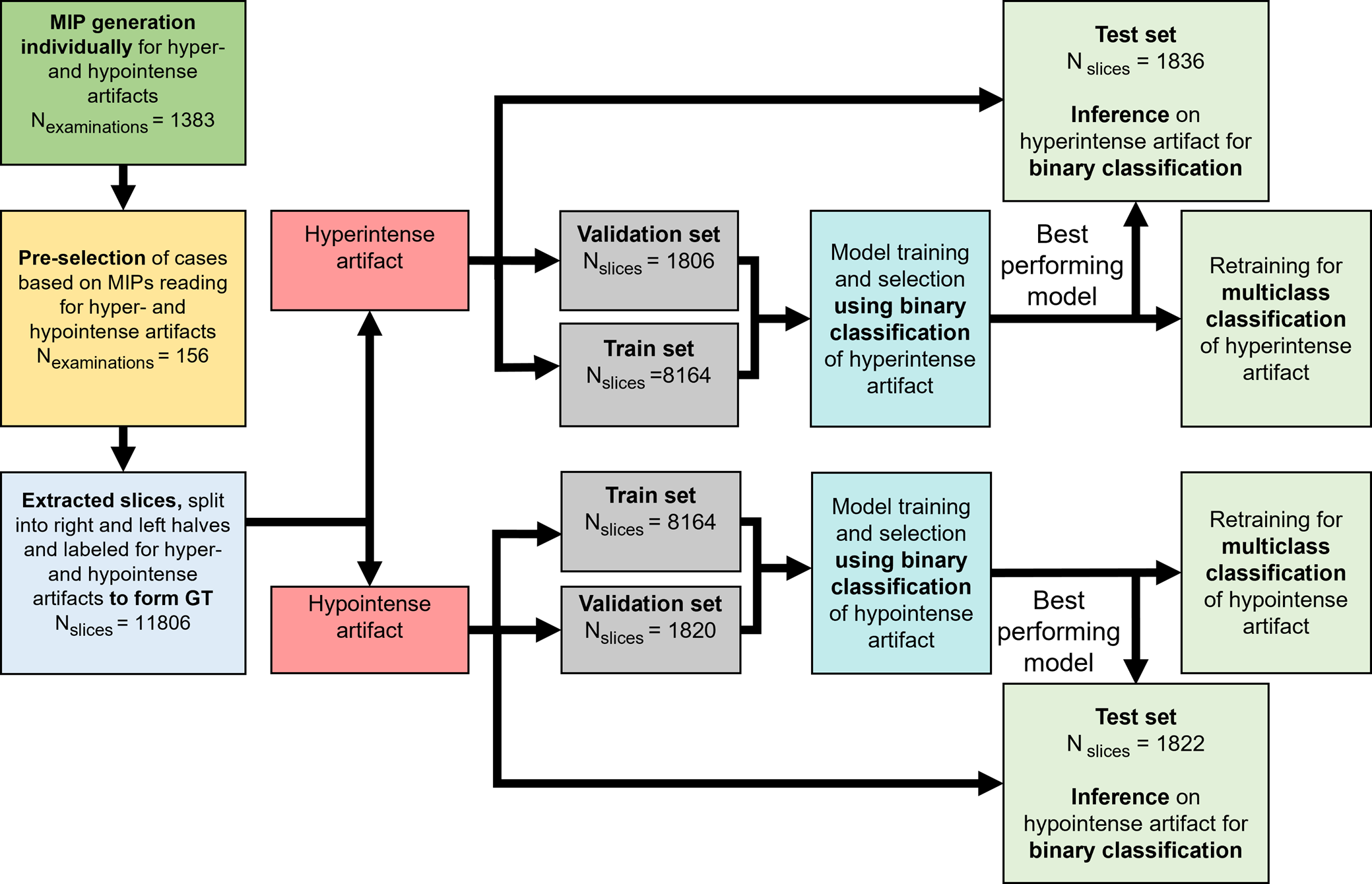}
\caption{Schematic overview of the study workflow, including maximum intensity projection (MIP)-based case pre-selection, slice-wise data generation to form ground truth (GT), data splitting, and model selection using DenseNet121, ResNet18, and SEResNet50 architectures on the validation set for binary classification of hyper- and hypointense artifacts. The best performing model was used for inference on the test set for binary classification and subsequently retrained for multiclass classification of hyper- and hypointense artifacts.}
\label{fig:flowchart}
\end{figure}

This study applied a systematic workflow as illustrated in the workflow diagram shown in Figure \ref{fig:flowchart}.

\subsection{Pre-selection and Slice-wise Dataset Generation}
From the total cohort of n = 1383 cases, n = 156 cases were pre-selected to be included into the CNN training. This pre-selection was done based on qualitative artifact assessment performed by one reader (D.S., doctoral researcher > 3 years of experience in DWI MRI). Preselection was done using MIPs, which were chosen as preparatory groundwork because they preserve artifacts across slices while reducing data dimensionality, thus reducing the reading time \cite{bickelhaupt2016fast, bickelhaupt2017maximum, kang2017unenhanced}.
MIPs were generated separately for hyper- and hypointense artifacts for each case using an in-house developed Python script. The MIPs were computed along the head-to-feet direction, selecting the maximum intensity value for each (x, y) coordinate in the transverse plane. This procedure transformed the 3D DWI volumes into 2D MIPs, allowing to evaluate artifact presence and intensity of a whole volume on a 2D image. The procedure for creation of MIPs is described in Supplemental Chapter 1.

Artifact severity was rated by the reader on a 5-point Likert-like scale where 1 indicated no artifact, 2 minor artifacts without diagnostic relevance, 3 moderate artifacts with potential diagnostic impact, 4 pronounced artifacts, and 5 severe artifacts \cite{dickinson2013scoring}. Cases were included in the training cohort if they received scores of 4 or 5 for both hyper- and hypointense artifacts. Additionally, 20 cases each were selected where the hyperintense artifact score was 2 or 3 with a hypointense artifact score of 5, and where the hypointense artifact score was 2 or 3 with a hyperintense artifact score of 5. This selection criterion ensured that all selected cases contained at least moderate to severe artifacts in either category.

As artifact severity varied across slices within a volume, the slice-wise dataset contained both artifact-free and artifact-affected samples, enabling the model to learn a broad range of artifact representations.
The pre-selected dataset of n = 156 cases was subsequently converted into a slice-wise dataset to create the final datasets for labeling, followed by training, validation and testing the CNNs. From the 156 cases, a total of 5903 slices were generated. The slices were separated into right and left breasts by dividing the width of the slice by two, resulting in 11806 slices. The slices were rescaled from raw image intensities to the range of (0--255) to make them consistent across all samples and to make them compatible with JPEG image format. The processed slices were saved in JPEG format while retaining the original orientation. The generated images were then resized to (160 x 128) spatial size using scaling to maintain uniformity for training purposes.

\subsection{Data Labeling}
\begin{figure}[h!]
\centering
\includegraphics[scale=1]{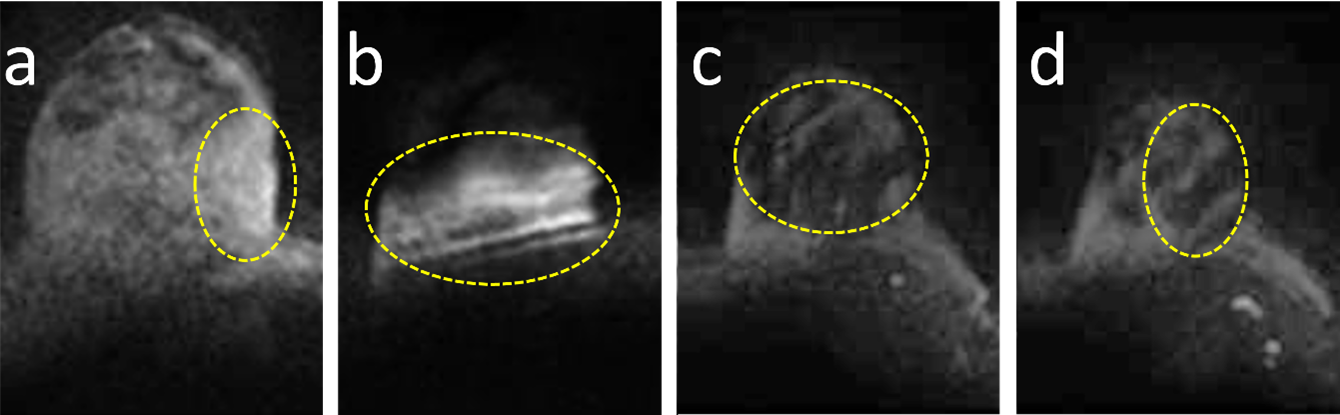}
\caption{Slices affected by hyper- and hypointense artifacts on DWI (b = 1500~s/mm$^2$): a) the enclosed region shows hyperintense artifact caused by surface coil flare; b) the enclosed region shows hyperintense artifact that is likely caused by skin folding; (c-d) enclosed regions depict hypointense artifacts.}
\label{fig:artifact img}
\end{figure}

For generating the ground truth (GT) for our slice-wise binary and multiclass training, the slices were annotated based on the presence of both hyper- and hypointense artifacts. Figures \ref{fig:artifact img}a and \ref{fig:artifact img}b illustrate hyperintense artifacts, whereas Figures \ref{fig:artifact img}c and \ref{fig:artifact img}d illustrate hypointense artifacts. 
Each dataset was annotated twice, once for hyperintense artifacts and once for hypointense artifacts. Data labeling to establish GT was performed by one reader (A.M., master’s student 6 months of experience in breast DWI) under the supervision of a board-certified radiologist (S.B., > 10 years of experience in breast MRI).

Artifact severity was rated on a 6-point Likert-like scale, where 1 indicated no artifact, 2 minor artifacts without diagnostic relevance, 3 moderate artifacts with potential diagnostic impact, 4 pronounced artifacts, 5 severe artifacts, and 6 ambiguous cases. All slices with a score of 6 were re-evaluated by a board-certified radiologist (S.B.) to resolve the discrepancies. 

To validate the reliability of the established GT, a group of two control readers (L.B. (V1), doctoral researcher 1.5 years of experience in breast DWI; I.H.  (V2), medical student 10 months of experience in radiology) with similar experience evaluated the holdout test set of hyper- and hypointense artifact. Both readers were instructed on how to read artifacts on demonstration cases in a separate session under the guidance of a board-certified radiologist (S.B.).

\subsection{Data Splitting}
The n = 11806 labeled slices were randomly split into training, validation, and test sets using a stratified split of 70\%, 15\%, and 15\%, respectively. This resulted in 8,164 slices in the training set, 1,806 slices in the validation set, and 1,836 slices in the holdout test set for the hyperintense artifact classification. For the hypointense artifact classification, the training set contained 8,164 slices, the validation set contained 1,820 slices, and the holdout test set contained 1,822 slices. Stratification was performed based on artifact scores to preserve the overall score distribution across all sets. To avoid data leakage, the split was conducted at the case level, ensuring that all slices from a given case were assigned exclusively to one subset. The number of slices differed in each split of hyper- and hypointense artifact due to the stratification and data leakage criteria applied.

\subsection{Data Preparation: Masking, Preprocessing, and Augmentation}

Breast masks were applied slice-wise to the images by element-wise multiplication, restricting network training to breast tissue and excluding background and non-breast anatomical structures. The original binary masks were generated from T1-weighted volumes using an in-house algorithm that applied mean thresholding to MIPs of adjacent slices along the z-axis, followed by binary dilation and binary closing operation to obtain homogeneous masks as described by Liebert et al. \cite{liebert2024feasibility}.

Masks were initially created for T1-weighted images, as direct mask generation on high b-value DWI is unreliable due to poor signal-to-noise ratio. These masks were therefore reused and resampled to match the spatial resolution of the DWI data.
Residual thoracic regions were removed by automatically cropping the masks around the sternum using an in-house detection algorithm. The final 3D masks were then applied slice-wise, yielding breast-only images for neural network training.

\subsection{Deep Learning: Training and Setup}
The binary classification was carried out for hyper- and hypointense artifact intensities using DenseNet121, ResNet18, and SEResNet50 models \cite{huang2017densely, he2016deep, hu2018squeeze}. Binary classification was chosen for practical use in a clinical setting as a tool for quality control capable of providing feedback and maybe even recommending to rescan the patient. 

For binary classification, artifact scores 1 and 2 were grouped as non-significant artifacts/no artifact (Binary Class 0) and scores 3–5 were grouped as potentially significant artifacts (Binary Class 1). 

All CNNs: were evaluated as a hyperparameter in the binary classification task on the validation set. Based on this hyperparameter on the validation set of hyper- and hypointense artifacts, the best-performing model was selected for inference on the holdout test set. The best-performing model was also trained separately for multiclass classification of hyper- and hypointense artifacts. This strategy was implemented, adhering to the general machine learning practices mentioned by Raschka \cite{raschka2018model}. 

Multiclass classification was implemented primarily for research purposes, allowing for a more detailed assessment of artifact severity. This approach offers flexibility in defining artifact severity threshold for clinical decision making and in evaluating the artifact prevalence and intensity across datasets. The original artifact scores (1--5) were retained as separate classes during all multiclass experiments.

All networks: DenseNet121, ResNet18, and SEResNet50 were implemented using MONAI with PyTorch Lightning \cite{cardoso2022monai, paszke2019pytorch, william_falcon_2020_3828935}. Training was conducted on NVIDIA RTX 2080 GPUs (8GB memory). Data loading was performed using MONAI dataloaders. 
All experiments used data balancing to mitigate class imbalance. During training of binary and multiclass experiments, a weighted random sampler was applied based on class weights, thereby increasing the sampling frequency of underrepresented classes. Models were optimized using the Adam optimizer with a cross-entropy loss function and a batch size of 32. Slices were first min-max normalized to the range [0,1] and data augmentation was performed using MONAI library with PyTorch as backend. Augmentations included random rotation (± 12°, probability $p = 0.5$) and random flipping along the x-axis and y-axis (probability $p = 0.5$). To account for differences between artifact types and classification settings, learning rates were tuned separately for each experiment. The final learning rates applied in training ranged from $4 \times 10^{-6}$ to $9 \times 10^{-5}$, depending on the model architecture and classification task (for details see, Table \ref{summary}). These learning rates were selected by trial and error based on the validation loss. Training was run for a maximum of 200 epochs and a minimum of 40 epochs, with early stopping applied based on validation loss (patience = 10 epochs). SoftMax was used for class probability estimation.

\begin{table}[h!]
\centering
\caption{Summary of experiments conducted, showing the network architectures, configurations, and learning rates used for classifying hyper- and hypointense artifacts.}
\label{summary}
\begin{tabular}{l l l}
\hline
\textbf{Network} & \textbf{Classification} & \textbf{Learning Rate} \\
\hline
\multicolumn{3}{c}{\textbf{Hyperintense Artifact}} \\
\hline
DenseNet121   & Binary      & $8 \times 10^{-6}$ \\
ResNet18      & Binary      & $4 \times 10^{-6}$ \\
SEResNet50    & Binary      & $8 \times 10^{-6}$ \\
DenseNet121   & Multiclass  & $8 \times 10^{-6}$ \\
\hline
\multicolumn{3}{c}{\textbf{Hypointense Artifact}} \\
\hline
DenseNet121   & Binary      & $9 \times 10^{-5}$ \\
ResNet18      & Binary      & $1 \times 10^{-5}$ \\
SEResNet50    & Binary      & $5 \times 10^{-5}$ \\
DenseNet121   & Multiclass  & $9 \times 10^{-5}$ \\
\hline
\end{tabular}
\end{table}

\subsection{Visualization of Network Prediction}

To support qualitative evaluation and visualization of the artifact regions identified by the models, bounding boxes were created based on Grad-CAM heatmaps. Grad-CAM was used to visualize spatial regions influencing model predictions by generating heatmaps that represent the relative contribution of each region \cite{selvaraju2017grad}. The heatmaps, with values ranging from 0 (highest activation) to 1 (lowest activation), were thresholded at 0.2 to retain the top 20\% activated regions. A binary mask was created from the thresholded heatmap, and external contours of this mask were extracted. Bounding boxes were then generated from these contours to enclose the regions most influential in the model’s decision.

\subsection{Performance Evaluation}
Performance evaluation for binary classification of hyper- and hypointense artifacts was first conducted on the validation set. After selecting the best-performing model, final performance evaluation was performed on the holdout test set. For multiclass classification, performance evaluation was conducted exclusively on the holdout test set.

For quantitative evaluation, the following performance metrics were employed: accuracy, precision, area under receiver operating characteristic curve (AUROC), area under precision recall curve (AUPRC), recall, and the confusion matrix. In the multiclass setting, AUROC was calculated using a one-vs-rest approach. For accuracy, precision, AUPRC and recall in the multiclass setting, weighted averages were used. AUROC, AUPRC and the confusion matrix were generated using sci-kit library, and accuracy, precision, and recall were computed using torchmetrics.

For the qualitative evaluation of the bounding boxes, a board-certified radiologist (T.T.N., 9 years of experience in breast MRI) assessed visually the bounding box position of 200 randomly selected true positive cases from the validation dataset across models. For final testing, bounding boxes were generated and evaluated for 200 randomly selected true positive cases from the holdout test set of the best-performing model.

Bounding boxes generated as described above were scored on a 5-point Likert-like scale with 1 indicating no overlap with the artifact, 2 a loose overlap with substantial inclusion of non-artifact regions, 3 artifacts generally enclosed but with noticeable boundary overreach or under-coverage, 4 good alignment with only minor boundary deviation or small amounts of adjacent tissue included, and 5 precise location of artifact.

The agreement between the best model predicted class in multiclass setting, GT, V1, and V2 was evaluated using Cohen’s kappa coefficient, separately for hyper- and hypointense artifacts on the holdout test set. Firstly, agreement between the best model predicted class, V1, and V2 was evaluated. Secondly, inter-reader agreement among GT, V1, and V2 was assessed. Agreement strength in all cases was interpreted according to the criteria proposed by Landis and Koch \cite{landis1977measurement}. The kappa coefficients were calculated with 95\% confidence interval (CI).

\section{Results}

\subsection{Reading Evaluation}

Data labeling the GT on the 11,806 slices, resulted in 19.3\% of slices being labeled to contain hyperintense artifacts (hyperintense artifacts labeled Classes 3--5), while 38.8\% were labeled to contain by hypointense artifacts (hypointense artifacts labeled Classes 3--5).

In the training set, the presence of artifacts was observed in 17.8\% (1455/8164) of slices. In the validation set, the presence of artifacts was observed in 23.1\% (418/1806) of slices, while in the holdout test set, 22.3\% (409/1836) of slices contained hyperintense artifacts.

For hypointense artifacts in the training set, 38.6\% (3149/8164) of slices were reported to be affected. The validation set showed a similar proportion with artifacts in 38.9\% (708/1820) of slices, while the holdout test set had slightly higher occurrence of 39.6\% (722/1822) slices.

\subsection{Hyperintense Artifact}

\subsubsection*{Binary Hyperintense Artifact Classification}
After evaluating on the validation set, DenseNet121 achieved the highest AUROC (0.89) and AUPRC (0.73), followed by ResNet18 (AUROC: 0.87, AUPRC: 0.72) and SEResNet50 (AUROC: 0.84, AUPRC: 0.66) (see, supplemental Table \ref{tab:ST2}). 

As DenseNet121 had better qualitative and quantitative performance on the validation set as compared to other networks (Supplemental Chapter 2, Supplemental Table \ref{tab:ST2}, Supplemental Figures \ref{fig: SF1} and \ref{fig: SF2}), inference of the network was done on the holdout test set for binary classification. Multiclass classification of hyperintense artifact was also carried out using DenseNet121.

\begin{figure}[h!]
\centering
\includegraphics[scale=1]{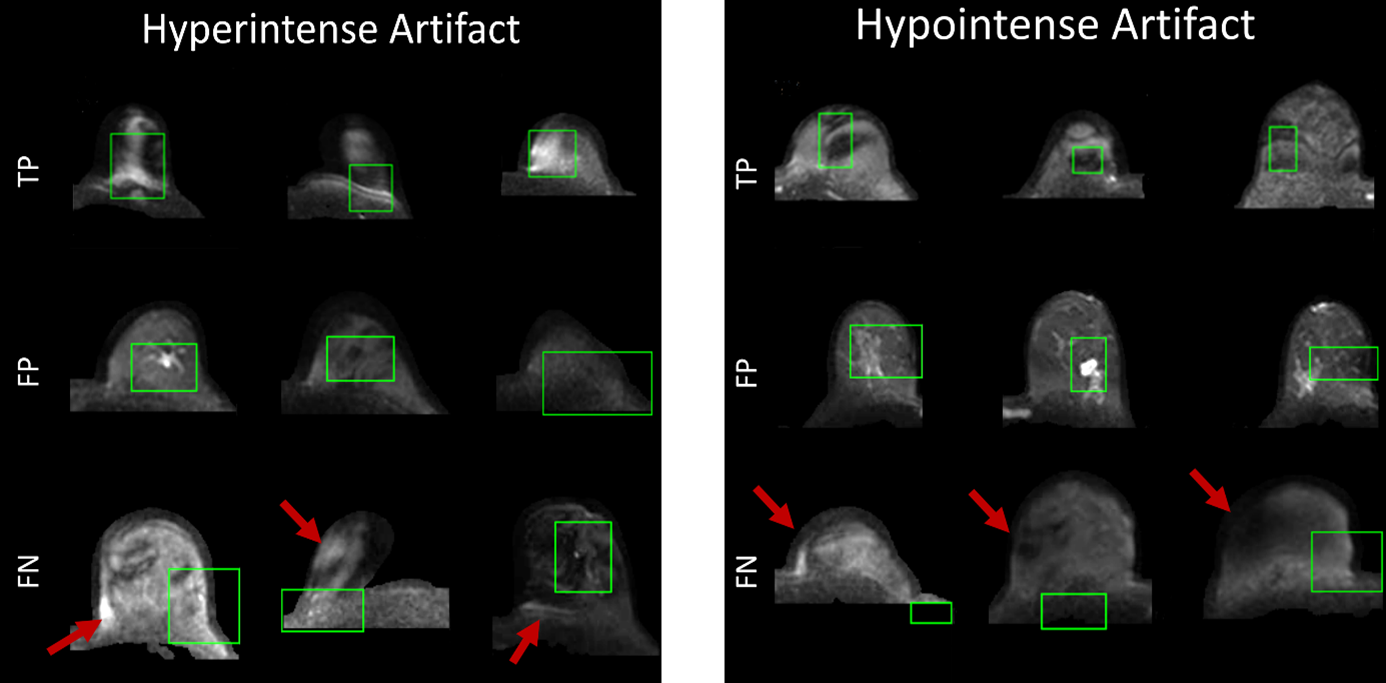}
\caption{True Positive (TP), False Positive (FP), and False Negative (FN) cases of bounding boxes predicted on masked high b-value (b = 1500~s/mm$^2$) DWI slices for binary hyper- and hypointense artifact classification on the holdout test set using DenseNet121. In FN cases, the red arrow shows the correct region of artifact.}
\label{fig:binary}
\end{figure}

The independent holdout test set of the hyperintense artifact detection training included a total of n = 24 cases, comprising 1836 slices. DenseNet121 achieved an AUROC of 0.92 and AUPRC of 0.77 on the holdout test set (see, Table \ref{tab:binary_test}). Bounding boxes generated from DenseNet121 predictions on the holdout test set for binary hyperintense artifact detection are shown in Figure \ref{fig:binary}.

Bounding box quality evaluated on 200 randomly selected slices from the holdout test cases on DenseNet121 resulted in, 9.5\% of the cases with a score of 5, 43.5\% of the cases with a score of 4, 21\% of the cases with a score of 3, 22.5\% of the cases with a score of 2, and 3.5\% of the cases with a score of 1. The mean score was determined to be 3.33 ± 1.04.

After evaluating DenseNet121 as the best performing network for binary hyperintense artifact classification on the validation set, the network was retrained for multiclass classification of hyperintense artifacts.

\begin{table}[h!]
\centering
\caption{Quantitative performance metrics (Accuracy, Precision, Recall, area under receiver operating characteristics curve (AUROC), and area under precision recall curve (AUPRC)) for binary hyper- and hypointense artifact classification using DenseNet121 individually on the holdout test set.}
\begin{tabular}{lccccc}
\hline
\textbf{Model} & \textbf{Accuracy} & \textbf{Precision} & \textbf{Recall} & \textbf{AUROC} & \textbf{AUPRC} \\
\hline
\multicolumn{6}{c}{\textbf{Hyperintense Artifacts (n = 1836 slices)}} \\
\hline
DenseNet121 & 0.84 & 0.60 & 0.82 & 0.92 & 0.77 \\
\hline
\multicolumn{6}{c}{\textbf{Hypointense Artifacts (n = 1822 slices)}} \\
\hline
DenseNet121 & 0.85 & 0.76 & 0.91 & 0.94 & 0.92 \\

\hline
\end{tabular}
\label{tab:binary_test}
\end{table}

\subsubsection*{Multiclass Hyperintense Artifact Classification}

\begin{figure}[htbp!]
\centering
\includegraphics[scale=0.9]{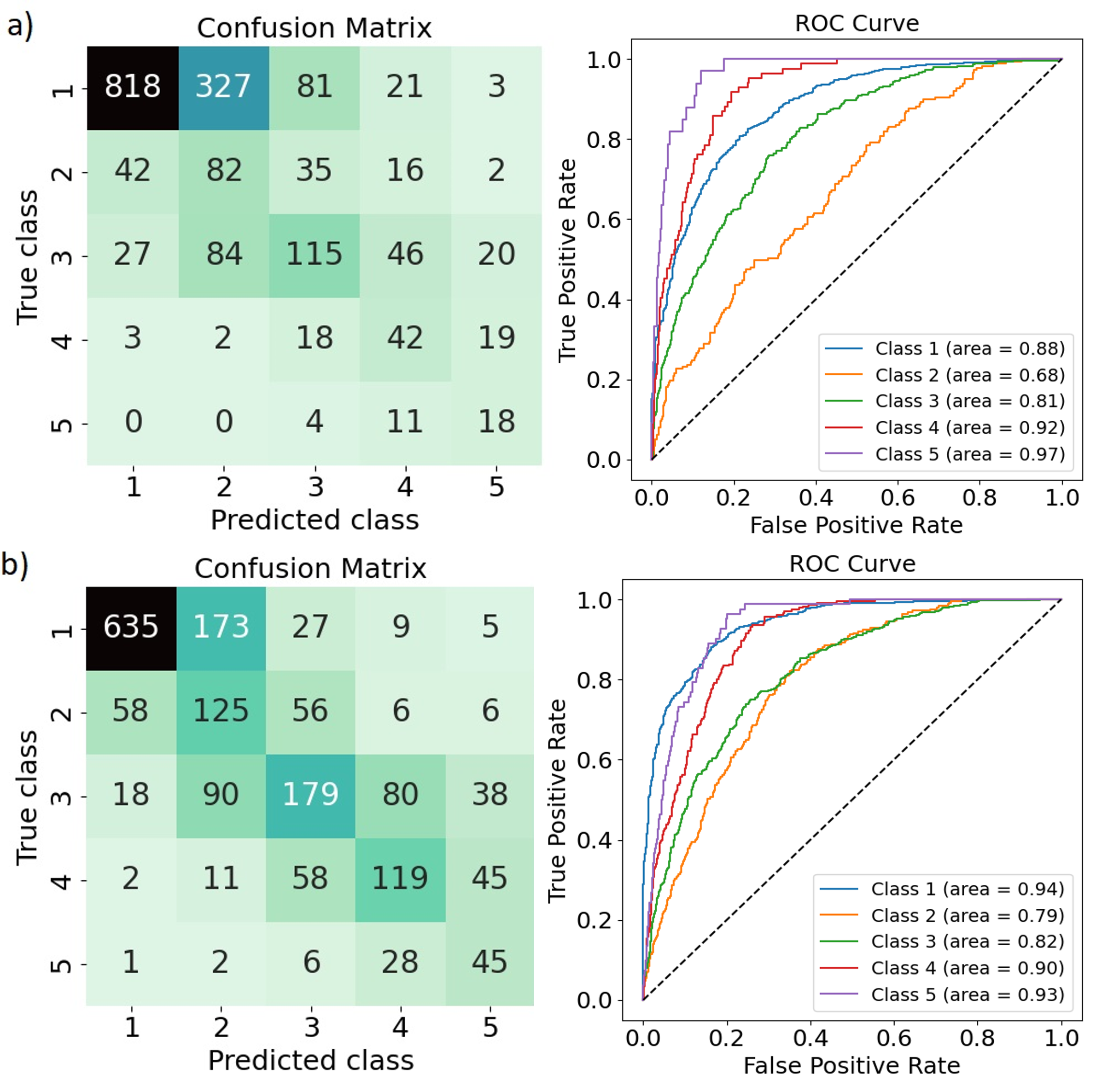}
\caption{Class-wise confusion matrices and area under receiver operating characteristics curves (AUROCs) illustrating the performance of the DenseNet121 model for multiclass classification of hyper- and hypointense artifacts on their respective holdout test sets. a) Confusion matrix and AUROC for hyperintense multiclass classification. b) Confusion matrix and AUROC for hypointense multiclass classification.}
\label{fig: multi plots}
\end{figure}

The multiclass evaluation was performed on the holdout test set, showing for hyperintense artifact classification 0.85 (AUROC) and 0.75 (AUPRC) (see, Table \ref{tab:multiclass_test}). The confusion matrix and the corresponding AUROC curve for each class are presented in Figure \ref{fig: multi plots}a, demonstrating strong class-wise performance, with AUROC values of 0.97 for Class 5 (severe artifact) and 0.88 for Class 1 (no artifact). Interestingly, DenseNet121 never misclassified Class 5 (extreme artifact affected class) slices as Class 1 (no artifact) or Class 2 (minor artifact) indicating that extreme artifacts were consistently distinguished from minimal to no artifacts.

\begin{table}[htbp!]
\centering
\caption{Weighted quantitative performance metrics (Accuracy, Precision, Recall, area under receiver operating characteristics curve (AUROC), and area under precision recall curve (AUPRC)) when predicted for multiclass hyper- and hypointense artifact classification using DenseNet121 on the respective holdout test sets.}
\begin{tabular}{lccccc}
\hline
\textbf{Model} & \textbf{Accuracy} & \textbf{Precision} & \textbf{Recall} & \textbf{AUROC} & \textbf{AUPRC} \\
\hline
\multicolumn{6}{c}{\textbf{Hyperintense Artifacts (n = 1836 slices)}} \\
\hline
DenseNet121 & 0.59 & 0.76 & 0.59 & 0.85 & 0.75 \\
\hline
\multicolumn{6}{c}{\textbf{Hypointense Artifacts (n = 1822 slices)}} \\
\hline
DenseNet121 & 0.61 & 0.68 & 0.61 & 0.88 & 0.69 \\
\hline
\end{tabular}
\label{tab:multiclass_test}
\end{table}

\FloatBarrier

\subsection{Hypointense Artifact}

\subsubsection*{Binary Hypointense Artifact Classification}

For binary hypointense artifact detection on the validation set, DenseNet121 achieved the highest quantitative performance, with an AUROC of 0.92 and an AUPRC of 0.88. This result was consistent with its superior performance in hyperintense artifact detection reported above, as compared to ResNet18 (AUROC = 0.92, AUPRC = 0.87) and SEResNet50 (AUROC = 0.90, AUPRC = 0.84). Quantitative performance metrics on the validation set for binary hypointense artifact classification are provided in Supplemental Chapter 2 (see, Supplemental Table \ref{tab:ST2}). 

Despite achieving superior quantitative performance, DenseNet121 received comparatively lower qualitative ratings during visual assessment of bounding boxes for binary hypointense artifacts on the validation set compared with ResNet18 and SEResNet50 (see, Supplemental Figure \ref{fig: SF2}). However, based on the consistently higher quantitative metrics, DenseNet121 was selected for inference on the holdout test set. Multiclass classification of hypointense artifact was also carried out using DenseNet121.

For binary hypointense artifact classification, DenseNet121 reached 0.94 (AUROC) and 0.92 (AUPRC) on the 1822 slices (see, Table \ref{tab:binary_test}). Figure \ref{fig:binary} shows the bounding boxes generated from DenseNet121 on the holdout test set for binary hypointense artifacts.

Bounding box quality evaluated on 200 randomly selected slices from the holdout test set on DenseNet121 resulted in, 0\% cases with a score of 5, 7\% cases with a score of 4, 61.5\% cases with a score of 3, 18\% cases with a score of 2, and only 13.5\% cases with a score of 1 yielding a mean score of 2.62 ± 0.81.

\subsubsection*{Multiclass Hypointense Artifact Classification}

The multiclass classification of hypointense artifact was evaluated on the holdout test set using DenseNet121 which achieved overall 0.88 (AUROC) and 0.69 (AUPRC) (see, Table \ref{tab:multiclass_test}). Strong class-wise performance, with AUROC values of 0.93 for Class 5 (severe artifact) and 0.94 for Class 1 (no artifact) was observed (see, Figure \ref{fig: multi plots}b).

\subsection{Reader--Model Agreement Analysis}

The reader--model agreement was assessed on the holdout test set using GT, V1, V2 labeling and the multiclass DenseNet121 predicted classes for both hyper- and hypointense artifacts. Reader--model and inter-reader agreement analyses were performed to check the agreement of the DenseNet121 predictions relative to human annotation variability. 

On the holdout test set for hyperintense artifact, GT and V1 exhibited fair agreement with the DenseNet121 predicted classes, while V2 demonstrated slight agreement with the DenseNet121 predicted classes. On the holdout test set for hypointense artifact, GT exhibited moderate agreement with DenseNet121 predicted classes, while V1 and V2 demonstrated fair agreement with the DenseNet121 predicted classes (see in Figure \ref{fig: kappa}). The Cohen’s kappa values with 95\% CI are summarized in Supplemental Table \ref{tab:ST3}.

\begin{figure}[htbp!]
\centering
\includegraphics[scale=0.8]{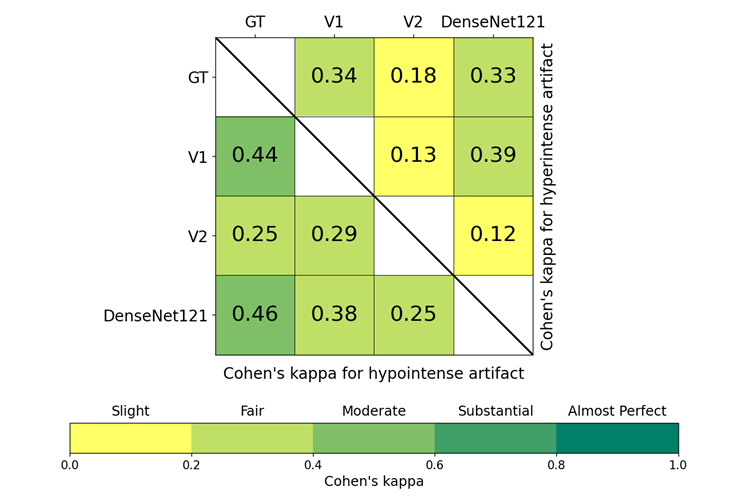}
\caption{The heatmap shows the agreement between ground truth (GT), Validator 1 (V1), and Validator 2 (V2) and the classes predicted by DenseNet121 on the holdout test set using Cohen's kappa score for hyper- and hypointense artifact obtained from multiclass classification.}
\label{fig: kappa}
\end{figure}

\section{Discussion}

Breast MRI increasingly incorporates DWI sequences, aiming at supporting lesion detection and characterization. High b-value DWI facilitates lesion depiction with the suppression of healthy breast parenchyma, however high b-value DWI can be prone to artifacts potentially impeding diagnostic assessment. Using a slice-wise breast MRI dataset and DWI acquisitions with a high b-value of 1500~s/mm$^2$, we demonstrated the ability of CNNs to detect and classify both hyper- and hypointense artifacts. 

Amongst the evaluated networks, DenseNet121 performed best based on quantitative and qualitative evaluations. This effect was observed both for hyperintense and hypointense artifacts. 

For hyperintense artifacts, DenseNet121 was able to reliably detect and classify artifacts in a binarized scenario, and it performed consistently well on the separate holdout test set, reaching an AUROC of 0.92, AUPRC of 0.77, and a recall of 0.82. This consistency showed that the model generalized well and did not seem prone to extensive overfitting to the training data. Compared to other neural networks implemented in this study, there was evidence of improved precision and AUROC on the holdout test set as shown in Table \ref{tab:binary_test}, indicating better handling of potential false positives and a stronger ability to differentiate between artifact and non-artifact cases on the diffusion-weighted images. 

For binary hypointense artifacts, DenseNet121 performed similarly well, with an AUROC of 0.94, AUPRC of 0.92, and accuracy of 0.85 on the holdout test set.

This observation was consistent as well when examining the multiclass classification of both hyper- and hypointense artifacts on the holdout test set. Indeed, the in-depth analysis of the experiments using a confusion matrix revealed that there were no instances for hyperintense artifacts where samples from Class 5 (most severe artifact) were misclassified as Class 1 (no artifact) or Class 2 (minor artifact) and only three cases for hypointense artifacts, highlighting the model’s strong ability to detect highly severe artifacts which can be clinically meaningful by enabling to balance in between sensitivity and specificity and not putting an overburden of false positive calls onto the technicians in order to achieve a reasonable sensitivity. Conformingly, the highest AUROC was commonly observed for Class 5, reflecting particularly robust identification of the most critical artifacts. Class-wise AUROC was still lower for Classes 2 and 3 for both hyper- and hypointense artifact on multiclass classification. However, intermediate classes of artifacts can pose a challenge to human readers as well. This rather mediocre performance on the “intermediate” artifact levels might be of less clinical importance since mostly strong artifacts might be prompting repeat acquisitions or switching to more robust acquisition techniques such as resolve DWI or alternating in between inversion recovery and spectral attenuated inversion recovery fat saturation \cite{le2025fat}.

A potential reason for DenseNet121’s superior performance among the three networks might be found in DenseNet121’s architectural advantage of preserving low-level spatial information through dense connectivity. For each layer, the feature maps of all the preceding layers are used as inputs, and their own feature maps are used as inputs for all the subsequent layers \cite{meng2022differentiation}. This may be particularly beneficial in detecting subtle intensity artifacts where localized features are critical. In contrast, ResNet18 with fewer layers may lack the depth to capture such nuanced representations, while SEResNet50 includes channel-wise attention and may not offer the same spatial connectivity advantages as DenseNet121. This performance trend aligned with the work of Kapsner et al., where DenseNet121 outperformed ResNet18 in detecting artifacts in dynamic contrast enhanced MIPs \cite{kapsner2022automated}. There have been no additional literature reports on the ability of DenseNet121 to differentiate artifacts on high b-value breast DWI slice-wise datasets.  

Detection of artifacts in imaging is of high diagnostic importance, especially in oncologic imaging. Lesions detected in breast MRI are commonly small and artifacts leading to a loss of signal or to hyperintense areas might cover significant finding. The previous work by Kapsner et al. indicated presence of artifacts on DWI MIPs in up to 37\% of the cases \cite{kapsner2023image}. While our study cannot be directly compared due to the preselection based on MIPs and the single slice approach, our data complements this assessment by indicating that in the case of artifacts in MIPs frequently many individual slices are affected in the individual examination. 

As opposed to the previous work, the current study investigated different classes of artifacts, which might enable more precise feedback in order to suggest mitigation strategies for the technicians. Indeed, the awareness about common causes can support technicians in selecting adequate artifact reduction steps \cite{noda2022primer}. 

The study had several limitations. Evaluating the multiclass model against GT, V1, and V2 resulted in only slight to moderate agreement (for details see, Figure \ref{fig: kappa}) in the granular class specific predictions, which meant that relevant mislabeling may still have occurred as labeling artifact strength for human readers is a subjective task. Subtle artifacts were sometimes interpreted differently by the reader, leading to label noise, which can affect model performance --- especially in borderline cases where artifact severity is difficult to classify. 

In all the results discussed above, we used DenseNet121 for inference on holdout test set in binary classification and training of multiclass experiments of hyper- and hypointense artifacts based on the hypothesis that a network good or better at classifying might also be good or better at detecting, but this is a limitation of the study, as using "classification as detection" does not necessarily guarantee optimal detection capability. In future, model training should explicitly be done for artifact detection using detection models like YOLO-based architectures or Faster R-CNN instead of inferring detection capability from classification performance \cite{redmon2016you, ren2016fasterrcnnrealtimeobject}. 

An important limitation of this work was the use of bounding boxes derived from Grad-CAM heatmaps as an approximate method for localizing artifact regions. Seeing that it does not always produce correct bounding boxes is rather a hint towards the network not always focusing on correct regions of the image when classifying. This approach provided an interpretable solution in the absence of precise GT bounding boxes. However, the bounding box lacked the accuracy and consistency of YOLO-based architectures or Faster R-CNN. Such approaches might have improved spatial localization, particularly in complex artifact regions. However, it should be noted that the identification of artifact position was not the primary goal of the work.

The models were trained solely on labels reflecting artifact severity. Future work could investigate whether incorporating additional clinical and technical metadata, such as BI-RADS category, scanner model, and magnetic field strength, may further improve performance. Scanner hardware and field strength, which were not included during training, are known to influence artifact appearance due to differences in acquisition protocols and system-specific characteristics \cite{kapsner2022automated}.

High b-value DWI (b = 1500~s/mm$^2$) was chosen because intensity artifacts are more conspicuous. Direct extension of the trained model to substantially lower b-values is not feasible due to marked differences in image contrast. Extension to moderately lower b-values (b = 750~s/mm$^2$), is more realistic but would require relabeling of the data and retraining of the model as intensity artifacts are often less visually distinct, which cannot be addressed through simple fine-tuning. Further, the data was not specifically chosen to include cases, in which artifacts actually obscured a malignant lesion, thus we cannot draw any conclusion about the clinical impact on improving lesion detection by mitigating the risks of artifacts. 

Finally, the dataset was limited in size and originated from a single center. The lack of a multi-center dataset due to the absence of a publicly available dataset with very high b-value DWI will restrict the generalizability of the model and its robustness to the scanner and patient population variability.
In conclusion, neural networks were able to classify both hyper- and hypointense artifacts in high b-value (b = 1500~s/mm$^2$) breast MRI on a single-slice level and indicate their presence using a bounding box to increase explainability. Further research is needed to understand how the detection of artifacts of different classes might best be translated into actionable items enabling technicians to counteract during the MRI examination.

\section*{Acknowledgements}
This work was partially funded by the Bavarian Ministry of Economic Affairs, Regional Development and Energy. The project is co-financed by the Polish National Agency for Academic Exchange within Polish Returns Programme.

\section*{Competing Interests}
The authors declare no competing interests.

\bibliographystyle{unsrt}  
\bibliography{references}  

\newpage

\section*{Supplementary Information}

\setcounter{table}{0}
\renewcommand{\tablename}{Supplemental Table}
\renewcommand{\thetable}{S\arabic{table}}
\renewcommand{\theHtable}{S\arabic{table}}

\setcounter{figure}{0}
\renewcommand{\figurename}{Supplemental Figure}
\renewcommand{\thefigure}{S\arabic{figure}}
\renewcommand{\theHfigure}{S\arabic{figure}}

\subsection*{Supplemental Chapter S1: Methods}
\label{SC1}

\begin{table}[h!]
\centering
\caption{Summary of diffusion weighted imaging (DWI) acquisition parameters including repetition time (TR), echo time (TE), inversion time (TI), matrix size and slice thickness for the dataset used in our study.}
\resizebox{\linewidth}{!}{
\begin{tabular}{lcccccc}
\hline
\textbf{Sequence} & \textbf{TR (ms)} & \textbf{TE (ms)} & \textbf{TI (ms)} & \textbf{Matrix Size} & \textbf{Slice Thickness (mm)} \\
\hline
DWI acquired by Skyra-Fit & 7030--8360 & 66 & 250 & 256--264 $\times$ 160--208 $\times$ 37--45 & 4 \\
DWI acquired by MAGNETOM VIDA & 5810--13440 & 66--80 & 220--250 & 256--264 $\times$ 160--208 $\times$ 37--45 & 4 \\
\hline
\end{tabular}}
\label{tab:ST1}
\end{table}

\subsubsection*{Creation of Maximum intensity projections (MIPs)}

For hyperintense artifacts, the 3D volumes were normalized, masked using binary masks, and projected directly using the MIP operation, which naturally enhances bright regions. Generation of binary masks is explained in Data Preparation: Masking, Preprocessing, and Augmentation subsection in the Materials and Methods chapter. The resulting MIPs were subsequently intensity-inverted ensuring hyperintense artifacts appear as dark structures in the final inverted MIP representations due to reader’s preferences to examine dark features on bright background.

For hypointense artifacts, which appear as dark regions in high b-value diffusion-weighted images, a direct MIP would suppress their visibility because the MIP emphasizes high-intensity voxels. To address this, the 3D diffusion weighted imaging (DWI) volumes were first normalized, intensity-inverted, and masked using binary masks before MIP generation.  The intensity inversion operation converted hypointense regions into bright regions, which allowed the MIP to highlight the hypointense artifacts.

\subsection*{Supplemental Chapter S2: Results}

\subsubsection*{Binary Classification of Hyper- and Hypointense Artifacts on the Validation Set}

\begin{figure}[htbp!]
\centering
\includegraphics[scale=1]{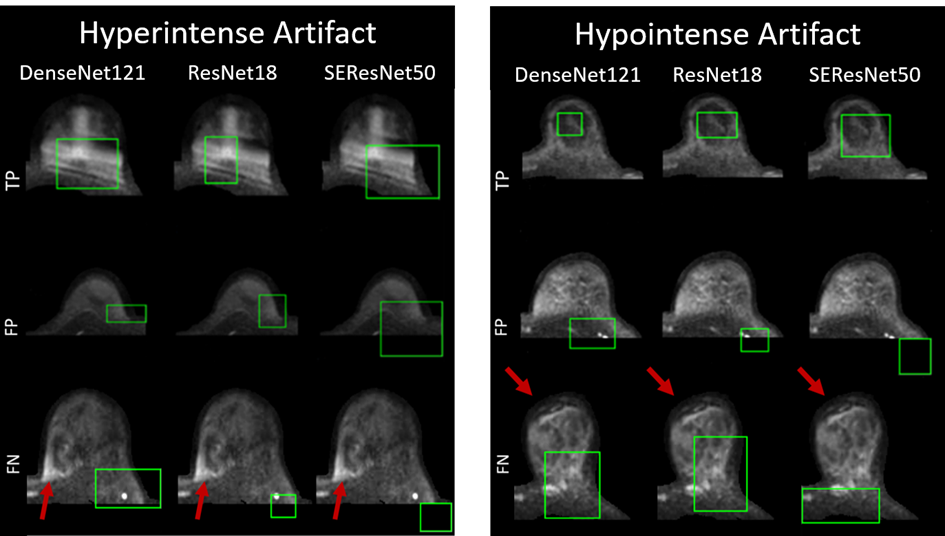}
\caption{True Positive (TP), False Positive (FP), and False Negative (FN) cases of bounding boxes predicted on masked high b-value (b = 1500~s/mm$^2$) DWI slices for binary hyper- and hypointense artifact classification on the validation set using DenseNet121, ResNet18, and SEResNet50. In FN cases, the red arrow shows the correct region of artifact.}
\label{fig: SF1}
\end{figure}

\begin{figure}[ht!]
\centering
\includegraphics[scale=1]{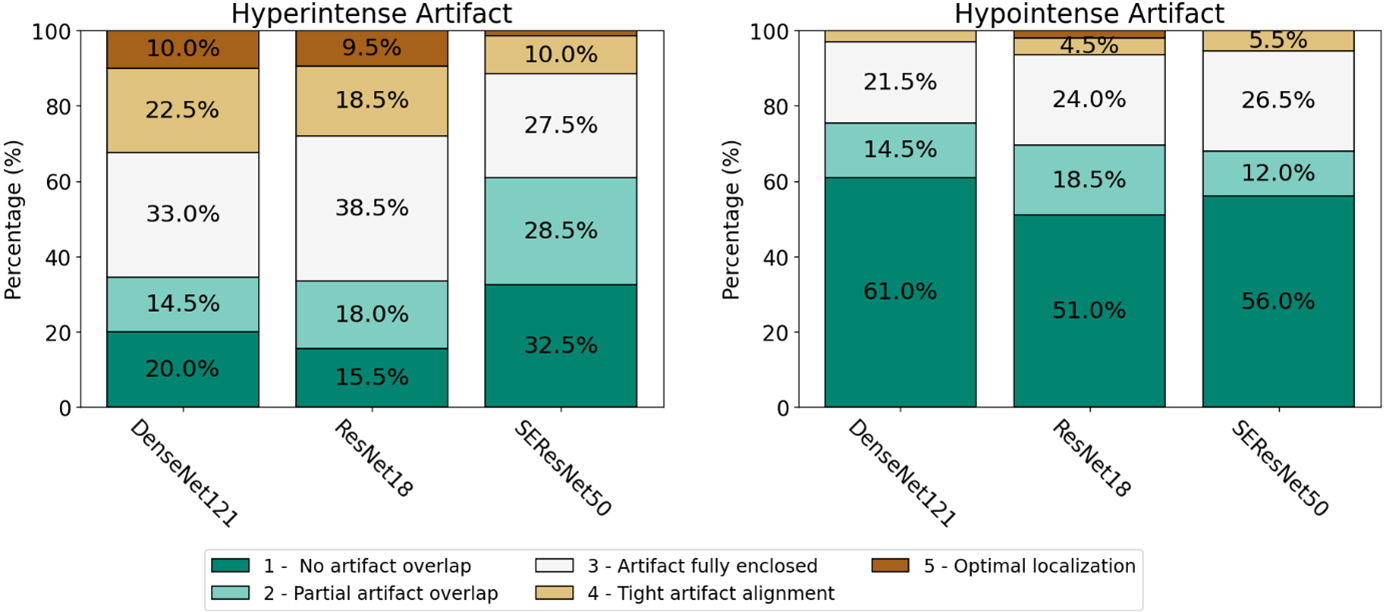}
\caption{Bounding box quality scores for scoring the ability to enclose an artifact on a 5-point Likert-like scale for DenseNet121, ResNet18, and SEResNet50 are shown in percentages on n = 200 validation set hyper- and hypointense artifact binary classification cases. }
\label{fig: SF2}
\end{figure}

On the validation set, DenseNet121 performed better as compared to ResNet18 and SEResNet50 in terms of detection accuracy and localization of hyperintense artifacts. Although ResNet18 showed comparable performance for hyperintense artifact detection. For hypointense artifact detection, DenseNet121 showed slightly inferior qualitative performance as compared to ResNet18 and SEResNet50, as demonstrated in both Supplemental Figure \ref{fig: SF1} and Supplemental Figure \ref{fig: SF2}.

Bounding box quality was evaluated on 200 randomly selected cases on the hyperintense validation set for DenseNet121, SEResNet50, and ResNet18. DenseNet121 and ResNet18 demonstrated the most consistent high-quality performance with mean quality score of 2.88 ± 1.25 for DenseNet121, 2.88 ± 1.17 for ResNet18, whereas SEResNet50 was rated with a lower mean score of 2.19 ± 1.05 (see, Supplemental Figure \ref{fig: SF2}).

For the hypointense artifact classification on the validation set, bounding box quality was evaluated on 200 randomly selected cases for DenseNet121, SEResNet50, and ResNet18. In contrast, DenseNet121 had 0 \% cases rated as a score of 5 and achieved lower mean score of 1.66 ± 0.91 as compared to slightly higher mean score of ResNet18 (1.88 ± 1.05) and SEResNet50 which also had 0\% cases rated a score of 5 with a mean score of 1.82 ± 1.01 (refer, Supplemental Figure \ref{fig: SF2}). Examples for the model generated bounding boxes for both hyper- and hypointense artifact for binary classification on the validation set are shown in Supplemental Figure \ref{fig: SF1}.

\begin{table}[h!]
\centering
\caption{The table shows the quantitative performance metrics (Accuracy, Precision, Recall, area under receiver operating characteristics curve (AUROC), and area under precision recall curve (AUPRC)) for binary hyper- and hypointense artifact classification using last-epoch checkpoints of DenseNet121, ResNet18, and SEResNet50 on the validation set.}
\begin{tabular}{lccccc}
\hline
\textbf{Model} & \textbf{Accuracy} & \textbf{Precision} & \textbf{Recall} & \textbf{AUROC} & \textbf{AUPRC} \\
\hline
\multicolumn{6}{c}{\textbf{Hyperintense Artifacts (n = 1806 slices)}} \\
\hline
DenseNet121 & 0.81 & 0.58 & 0.81 & 0.89 & 0.73 \\
ResNet18    & 0.80 & 0.56 & 0.79 & 0.87 & 0.72 \\
SEResNet50  & 0.79 & 0.55 & 0.74 & 0.84 & 0.66 \\
\hline
\multicolumn{6}{c}{\textbf{Hypointense Artifacts (n = 1820 slices)}} \\
\hline
DenseNet121 & 0.82 & 0.72 & 0.88 & 0.92 & 0.88 \\
ResNet18    & 0.83 & 0.75 & 0.85 & 0.92 & 0.87 \\
SEResNet50  & 0.80 & 0.69 & 0.86 & 0.90 & 0.84 \\
\hline
\end{tabular}
\label{tab:ST2}
\end{table}

\clearpage
\begin{table}[ht!]
\centering
\caption{Cohen's kappa with 95\% confidence interval (CI) agreement between ground truth (GT), Validator 1 (V1), Validator 2 (V2), and DenseNet121 predicted classes for hyper- and hypointense artifacts on the holdout test set.}
\begin{tabular}{lcccc}
\hline
 & \textbf{GT} & \textbf{V1} & \textbf{V2} & \textbf{DenseNet121} \\
\hline
\multicolumn{5}{c}{\textbf{Hyperintense Artifacts}} \\
\hline
GT & x & 0.34 (95\% CI: 0.31, 0.37) & 0.18 (95\% CI: 0.15, 0.21) & 0.33 (95\% CI: 0.3, 0.36) \\
V1&x & x & 0.13 (95\% CI: 0.11, 0.16) & 0.39 (95\% CI: 0.36, 0.42) \\
V2 & x & x & x & 0.12 (95\% CI: 0.09, 0.15) \\
\hline
\multicolumn{5}{c}{\textbf{Hypointense Artifacts}} \\
\hline
GT & x & 0.44 (95\% CI: 0.42, 0.47) & 0.25 (95\% CI: 0.22, 0.28) & 0.46 (95\% CI: 0.43, 0.48) \\
V1 & x & x & 0.29 (95\% CI: 0.26, 0.32) & 0.38 (95\% CI: 0.35, 0.41) \\
V2 & x & x & x & 0.25 (95\% CI: 0.22, 0.28) \\
\hline
\end{tabular}
\label{tab:ST3}
\end{table}

\end{document}